\documentclass[10pt,journal,compsoc]{IEEEtran}

\ifCLASSOPTIONcompsoc
  \usepackage[nocompress]{cite}
\else
  \usepackage{cite}
\fi
\usepackage{multirow}
\usepackage{amsmath}
\usepackage{amssymb}
\usepackage{graphicx}
\usepackage{array}
\usepackage{multirow}
\usepackage[thinlines]{easytable}
\usepackage{hhline}
\usepackage{bm}
\usepackage{color}
\usepackage{tabulary}
\usepackage{gensymb}
\usepackage{booktabs,caption}
\usepackage[flushleft]{threeparttable}
\usepackage{cite}
\newcolumntype{C}[1]{>{\centering\arraybackslash$}p{#1}<{$}}
\ifCLASSINFOpdf
\else
\fi
\begin{document}
%
\title{CNN-based RGB-D Salient Object Detection: Learn, Select and Fuse}
%
%
%
%

\author{Hao~Chen
        and~Youfu~Li,~\IEEEmembership{Senior~Member,~IEEE}
\IEEEcompsocitemizethanks{\IEEEcompsocthanksitem Hao Chen and Youfu Li are with the Department of Mechanical Engineering, City University of Hong Kong.
	\protect\\
E-mail: meyfli@cityu.edu.hk (Youfu Li is the corresponding author)
}
\thanks{Manuscript received April 19, 2005; revised August 26, 2015.}}

%
%

\markboth{Journal of \LaTeX\ Class Files,~Vol.~14, No.~8, August~2015}%
{Shell \MakeLowercase{\textit{et al.}}: Bare Advanced Demo of IEEEtran.cls for IEEE Computer Society Journals}
%



\IEEEtitleabstractindextext{%
\begin{abstract}
The goal of this work is to present a systematic solution for RGB-D salient object detection, which addresses the following three aspects with a unified framework: modal-specific representation learning, complementary cue selection and cross-modal complement fusion. To learn discriminative modal-specific features, we propose a hierarchical cross-modal distillation scheme, in which the well-learned source modality provides supervisory signals to facilitate the learning process for the new modality. To better extract the complementary cues, we formulate a residual function to incorporate complements from the paired modality adaptively. Furthermore, a top-down fusion structure is constructed for sufficient cross-modal interactions and cross-level transmissions. The experimental results demonstrate the effectiveness of the proposed cross-modal distillation scheme in zero-shot saliency detection and pre-training on a new modality, as well as the advantages in selecting and fusing cross-modal/cross-level complements.
\end{abstract}

}

\maketitle

\IEEEdisplaynontitleabstractindextext

%
\IEEEpeerreviewmaketitle

\ifCLASSOPTIONcompsoc
\IEEEraisesectionheading{\section{Introduction}\label{sec:introduction}}
\else
\section{Introduction}
\label{sec:introduction}
\fi

%
%
%
%
\IEEEPARstart{T}{he} availability of depth sensors (e.g., in Microsoft Kinect and Intel RealSense) allows the RGB-based computer vision systems with more accurate and robust performance, hence nurturing a wide range of applications \cite{1}. Complementary to the RGB data, the synchronized depth information carries additional geometry cues, which are immune to appearance changes, illumination varying and subtle background movements. The joint inference with RGB and depth information could benefit various computer vision tasks \cite{2,3}. A good example is the salient object detection \cite{4} of identifying the most visually attractive object/objects in a scene, which has been widely applied in image retrieval \cite{5} and object tracking \cite{6}. The RGB-based methods are very likely to fail when the salient object and background present similar appearance\cite{7,8,9,10}. From another perspective, the corresponding depth map, which supplies auxiliary saliency cues, opens up a new opportunity to solve this challenge.
\begin{figure*}[!t]
	\centering
	\includegraphics[height=0.31\textwidth, width=1\textwidth]{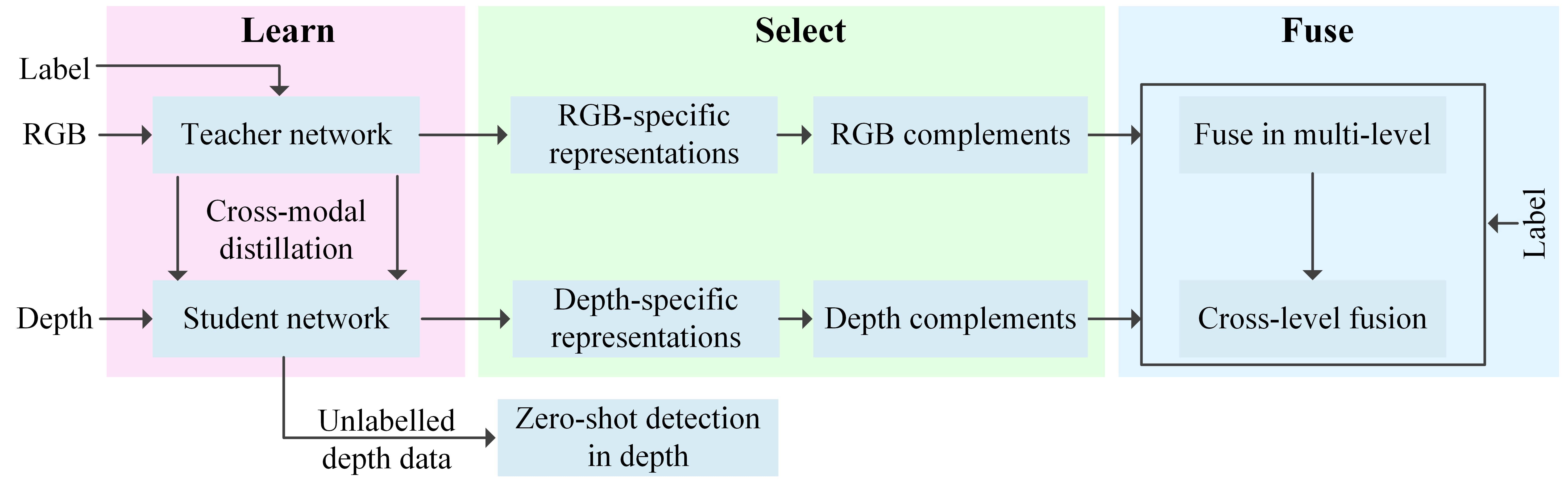}
	\setlength{\belowcaptionskip}{1pt}
	\caption{Our pipeline for RGB-D salient object detection.}
	\label{fig1}
\end{figure*}

\begin{figure*}[!t]
	\centering
	\includegraphics[height=0.5\textwidth, width=1\textwidth]{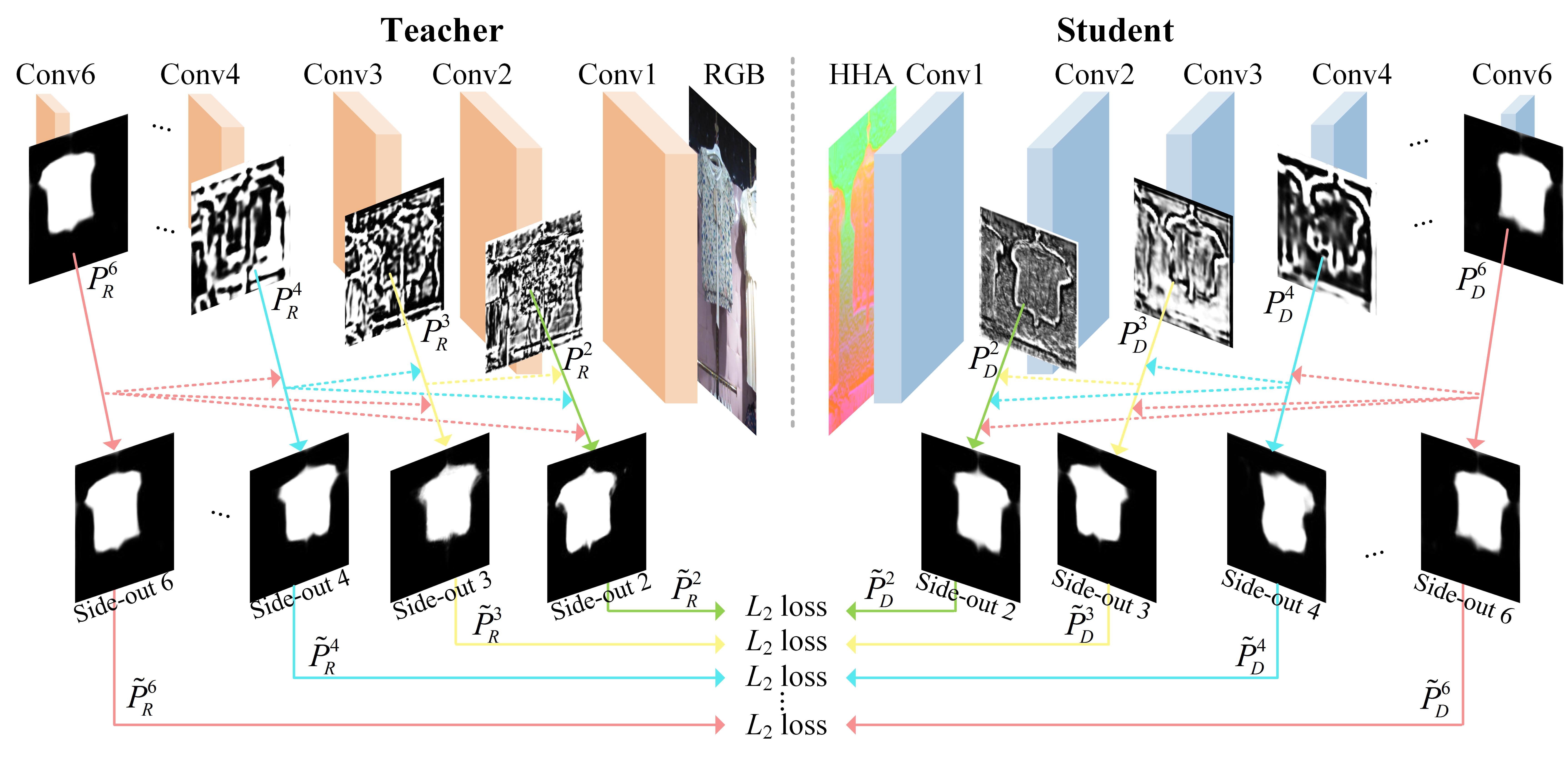}
	\setlength{\belowcaptionskip}{1pt}
	\caption{The architecture of the hierarchical cross-modal distillation network. Adaptation layers in each level are omitted for simplification. When training the teacher network, the L2 loss in each level is replaced with the cross-entropy loss between the side output and the ground-truth mask. For the depth data, we follow the previous approach \cite{43} to encode it as 3-channel HHA (horizontal, aboveground height and surface normal angle) representations.}
	\label{fig2}
\end{figure*}
\par By fusing the RGB and depth data, a rich amount of algorithms \cite{11,48,49,50,12,13,14,15,16,17,18,19,20,22,23} have been proposed for RGB-D salient object detection. Some previous works \cite{13,17,19,20} focus on crafting RGB-D features with prior knowledge, as the salient object tends to pop-out its surroundings. However, these nontrivial assumptions cannot be well generalized to all contexts. Another line of works \cite{4,14,15,16} infer saliency from each modality separately and then solve the multi-modal fusion problem by straightforward combination schemes. However, the cross-modal complements are not well integrated for better representations. Recently, the success of deep learning techniques \cite{21} in various computer vision tasks motivates more researchers to design RGB-D systems based on deep learning tools. A popular architecture is the ``two-stream" Convolutional Neural Network (CNN) \cite{22,23,24,25,26,27}, in which the paired RGB and depth images work independently and then aggregate in an early or late stage. In these networks, the depth stream is typically trained from scratch or initialized with the well-trained RGB CNN. Nonetheless, these training schemes typically end with insufficient depth-specific learning due to the scarcity of the labeled depth data. 
\par  Without carefully selecting the real complementary cues, the direct combination strategy in previous two-stream networks is also confronted with ambiguous and uninformative fusion. Moreover, with a single fusion layer as done in \cite{22,23,24,27}, it is unlikely to explore both the contextual and spatial cross-modal complementarities existed in multiple levels. Thus, the systematic solution for understanding RGB-D data still remains as an open issue. As illustrated in Fig. 1, we argue that an ideal RGB-D fusion system should successfully achieve the following three goals:
\par (1) \textbf{Learn}: In some scenarios, the specialists in one modality (e.g., geometry cues in the depth map) are missing in its counterpart (i.e., the RGB image). Accordingly, an informative RGB-D combination firstly calls for carefully extracting discriminative modal-specific features from each modality. Otherwise, knowledge from one modality may not assist and even mislead the inference for its counterpart. However, we are often confronted with an imbalanced amount of labeled data prepared for each modality. Thus, the challenge lies on how to learn rich modal-specific representations from the new modality with limited labeled data.
\par (2) \textbf{Select}:  An informative multi-modal fusion process should be attentive to the real complementary components. This awareness mechanism enables the cross-modal fusion to select complementary representations and ignore the redundant ones. 
\par (3) \textbf{Fuse}: The last step is to fuse the selected cross-modal information sufficiently. The complementarities between RGB and depth data exist in both high-level contexts and low-level spatial details. Consequently, a sufficient RGB-D fusion process is in demand to associate both the low-level and high-level cross-modal features for joint decision.
\par Considering the unavailability of large-scale labeled data in the depth modality, we leverage structured knowledge provided by the source modality (i.e., RGB) to aid the learning of the new modality. Specifically, we use the side outputs of the source modality (i.e., RGB) as supervision to learn the target modality (i.e., depth). We term our scheme \emph{\textbf{hierarchical cross-modal distillation}}, which eschews the reliance on saliency ground-truths in the new modality. 
\par To render an effective fusion process, we explicitly encode the cross-modal complements with the residual function and the goal of selecting cross-modal features is formulated as asymptotically approximating the residual. Different from the direct concatenation of multi-modal features, such a cross-modal residual connection is more likely to expose the desired complementarity.
\par Concerning sufficient multi-modal fusion, we adopt a top-down fusion manner, in which cross-modal features are combined in each level and the integrated RGB-D representations, in turn, guide the inference of shallower layers. The resulting network demonstrates rich multi-level RGB-D representations for joint inference and consequently, the saliency map quality is improved progressively from coarse to fine.
\par Our preliminary studies \cite{12} discussed the above-mentioned ``Select" and ``Fuse" problems. However, the problem ``Learn" remains under-studied. In this work, we extend \cite{12} by investigating the problem ``Learn" and propose the hierarchical cross-modal distillation method.
\par In summary, this work has the following five contributions:
\par (1) We systematically analyze the key issues in interpreting RGB-D data, which guides the system design process.
\par (2) We propose a cross-modal distillation scheme, which allows zero-shot detection or favors better learning of new modalities with limited labeled data. 
\par (3) The residual function is designed to explicitly capture the cross-modal complementarity.
\par (4) We propose a progressively top-down cross-modal cross-level fusion topology. Thus, the inference path comes to be aware of modal-specific and level-specific contributions.
\par (5) This work achieves state-of-the-art performance on three benchmarks consistently.

\begin{figure*}[!t]
	\centering
	\includegraphics[height=0.625\textwidth, width=1\textwidth]{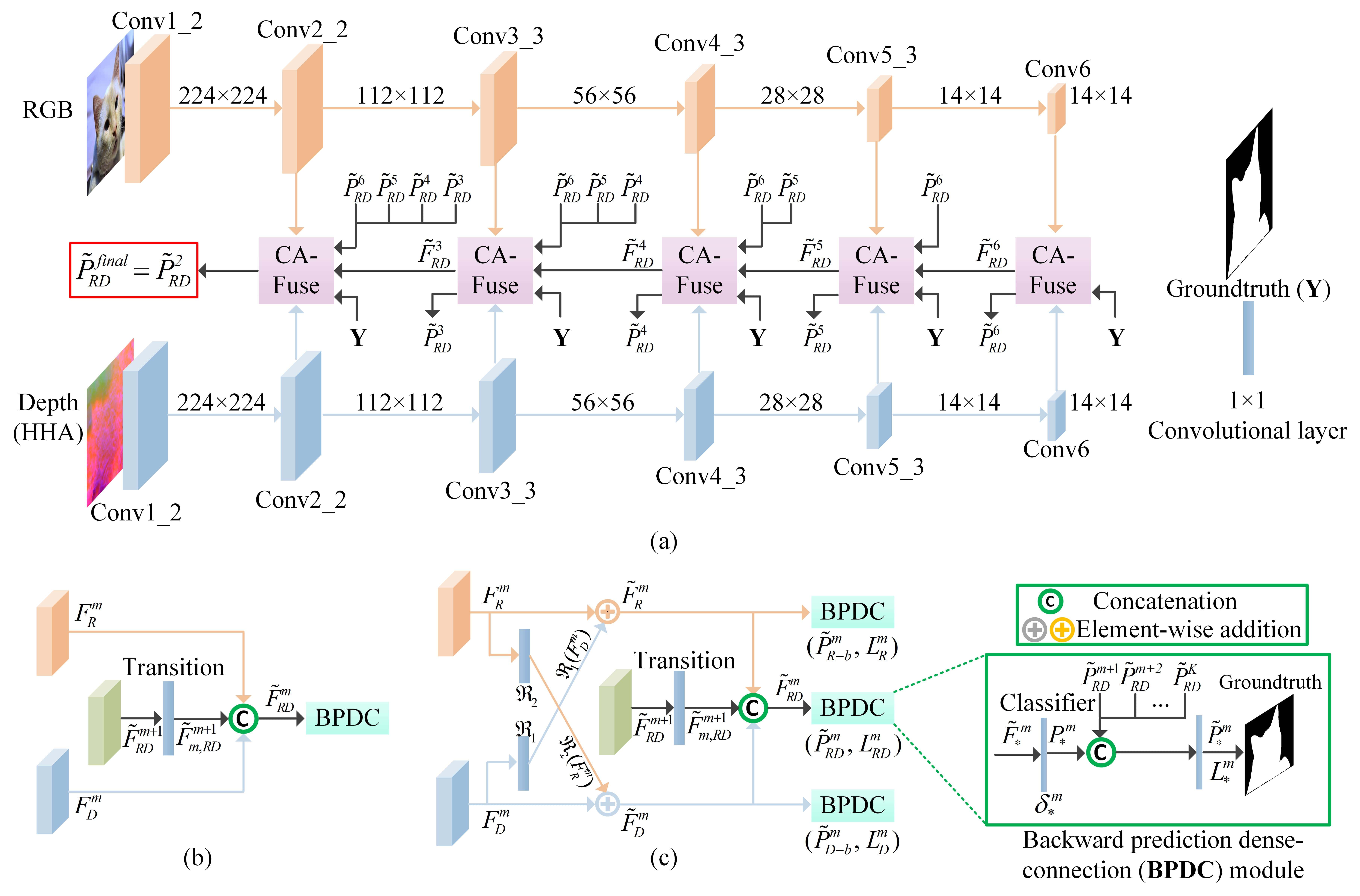}
	\setlength{\belowcaptionskip}{1pt}
	\caption{(a) The architecture of the RGB-D salient object detection network. (b) The details of direct concatenation of cross-modal cross-level features without the cross-modal residual designs. (c) The details of the complementarity-aware fusion (CA-Fuse) block.}
	\label{fig3}
\end{figure*}

 
\section{Related work}
\subsection{RGB-D Saliency Detection and Other RGB-D Systems}
A large body of earlier works focus on designing RGB-D features or combining unimodal predictions, which are termed as ``feature fusion'' and ``result fusion'' solutions respectively. A common wisdom in crafting depth-induced saliency cues is that human fixations prefer closer depth ranges. Based on this prior, Lang et al. \cite{14} use Gaussian Mixture Models to model the distribution of depth-induced saliency. This prior is useful but is easily confused by nearer backgrounds. On the other hand, two regions, sharing the same depth may be in different contexts and should be differentiated. Considering the scene structures, Ju et al. \cite{17} use relative depth instead of the absolute one for evaluation and propose the anisotropic center-surround difference for measurement. Desingh et al. \cite{15} adopt the global-contrast method \cite{9} used in the RGB-induced saliency detection with depth values as inputs. A similar framework is also used in \cite{28}. Different from these global-contrast paradigms, Feng et al. \cite{13} propose to measure the distinction of one region in a local context. They design a local background enclosure feature, which estimates the proportion of the object popping out the background. Peng et al. \cite{4} then propose a hybrid framework that incorporates global-contrast and local-contrast strategies. To further enrich the representative ability of RGB-D data, Song et al. \cite{19} segment the RGB-D pair into superpixels with different scales. The features are then combined as multi-scale representations.
\par Despite the effectiveness of these handcrafted features, they lack high-level reasoning and suffer from limited generalization ability. To address this limitation, recent works resort to deep learning techniques. Qu et al. \cite{22} combine the low-level features from RGB and depth modalities as the joint input to train a CNN from scratch. Compared to the previous works based on handcrafted features, this method achieves encouraging performance gains. However, it may be difficult to fully leverage the power of CNNs by feeding the crafted features rather than the raw image pair as inputs. In contrast, Han et al. \cite{23} design a ``two-stream'' late fusion architecture, in which the RGB and depth images are learned separately and their deep representations are combined by a joint fully connected layer for collaborative decision. Compared to \cite{22}, \cite{23} achieves large improvement due to the better combination of high-level contexts. Despite this, the low-level cross-modal complements are underexplored in \cite{23} and the resulting saliency maps are severely blurred. In summary, both \cite{22} and \cite{23} fail to combine the low-level and high-level cross-modal complementarity simultaneously. Recently, Chen et al. \cite{29} propose a multi-branch fusion network with fully connected layers for global reasoning and dilated convolutional layers for capturing local details. The results from two branches are combined by direct summation. However, the network is not a fully convolutional one and fails to utilize the information from all layers for joint inference. 
\par Deep learning techniques especially CNNs are also popular solutions for other RGB-D systems. Among which, the ``two-stream'' late fusion architecture is likewise the most widely-used one. In \cite{24} and \cite{27}, the multi-modal fusion layer combines the decisions from RGB and depth by modeling their consistency and independency. More recent works \cite{25,30,31,32} also investigate the cross-modal complementarity in multiple levels. Although the community of CNN-based RGB-D systems has achieved encouraging improvements, a comprehensive analysis on the RGB-D fusion problem is lacking, which, in our view, will benefit future works on multi-modal systems or new unlabeled modalities a lot. 

\subsection{Cross-modal transfer}
The transfer learning community mainly solve the domain adaptation problem in the same modality \cite{33,34,35,36}. In \cite{33}, Hinton et al. use the final soft outputs of the large well-trained teacher network as targets of the small student network. Subsequent works \cite{34,35,36} extend this idea by encouraging the student to mimic the deep representations from the teacher. 
Our topic lies in the cross-modal transfer problem, which is more difficult due to the severe cross-modal discrepancy. Notable works include \cite{51,52,53,54,55,37}. \cite{52,53} aims at learning joint representation by mapping the features from different modalities to a shared feature space. \cite{51} learns a mapping from the source modality to the unlabelled new ones to hallucinate modalities, while \cite{54,55} design a hallucinate network to distill depth features. Gupta et al. \cite{37} generalize the idea in \cite{34,35,36} to the cross-modal domain. However, due to the cross-modal discrepancy, a considerable part of source feature maps (e.g., texture and color changes in RGB images) are inaccessible for the target modality (i.e., depth). Hence, it is too strict for the new modality to mimic the high-dimensionality features from the teacher. These RGB-specific features provide uninformative and even negative supervision for the student. As a result, the student network is hard to converge especially when it is deep. In this work, the goal of the student network is relaxed to mimic the side outputs from each level in the teacher network. 
\section{The Proposed Method}
The proposed model can fulfill zero-shot detection in new unlabeled modalities as well as multi-modal joint inference. We address the three learning objectives with following stages: the teacher network with RGB images and ground-truth masks; the hierarchical cross-modal distillation network with un-annotated RGB-D pairs; and the multi-modal fusion network for RGB-D saliency detection with RGB-D pairs and ground-truths. In the following sections, we will follow the training sequence to introduce each network and discuss how the proposed solution behaves to learn, select and fuse cross-modal complementarity.
\subsection{Hierarchical Cross-modal Distillation}
For the cross-modal transfer learning, suitable supervisory signals should be customized for the student network. If a excessively strict constraint is set, it turns out to be difficult to ensure training convergence. In contrast, an over-relaxed constraint, such as appropriating the final class distributions, appears too weak to learn the shallow layers effectively. A well-balanced knowledge distillation method should provide sufficient supervisory signals while allows the exploration of specialists in a new modality. Intuitively, the features from two modalities, though discrepant, can make consistent inference for the same task. So our primitive choice is to use the inference from each level in the teacher network as supervision. However, as the observations in \cite{40}, the lower layers are more modal-specific and task-agnostic, while the deeper layers hold opposite characteristics. As a result, the shallow layers will hardly produce coherent inference by different modalities. Specifically, the shallower layers trained with RGB images are activated by texture/color changes, which are immune for the depth modality. We further consider that with the global guidance from the deep layers, the discrepancy between the combined side-outs in shallow layers across different modalities can be effectively reduced with respect to the individual inference in each level. 
Also, it is hard to optimize multiple levels from scratch jointly in a deep network. However, the progressive enhancement inferred from the teacher reveals level-specific contributions and cross-level collaborations, which are pretty informative supervisory signals for the student. 
To this end, we let the student progressively approach the side-outs from the teacher. We call it "hierarchical cross-modal distillation scheme". 
Such a design presents several distinguished advantages:
\par (a) Compared to the feature-based objective function, the relaxed inference-based supervisory signals allow more flexibility for the student to explore specialists. 
\par (b) These supervision signals decouple the joint learning of multiple layers and define level-specific optimization objectives for each level independently, which simplifies the training process for a deep student network.
\par (c) This method promotes a better transfer learning across modalities and even promises the feasibility of using deep CNNs to zero-shot detection on new visual modalities.
Given a new modality $\mathcal{M}_{D}^{X}$ with unlabeled training samples $\mathcal{X}^{D}$, our goal lies on learning model-specific features from $\mathcal{X}^{D}$ by transferring knowledge from a different modality $\mathcal{M}_{R}^{X}$ with large-scale labeled images.
\par Denote the ${K}$ layered representations $\Psi  = \{ \varphi _{R}^i,i = 1,...,K\}$, where $\varphi _{R}^i$ denotes the representation in the $i^{th}$ layer for the modality $\mathcal{M}_{R}^{X}$.
Based on $\varphi _{R}^i$, a reliable classifier $\delta _r^i$ is learned for level-specific inference $P_R^i = \delta _r^i(\varphi _{R}^i)$.  
\par Now, suppose we have a dataset $\mathcal{D}_{{r,d}}$ which contains sufficient un-annotated paired images from $\mathcal{M}_{R}^{X}$ and $\mathcal{M}_{D}^{X}$. 
We implement this idea by densely skip-connecting the inferences from the deeper layers to all lower layers to generate collaborative side outputs. As shown in Fig.2, the cross-modal distillation network contains two parts: (a) Unimodal cross-level connections (Note that the teacher and student share same cross-level connections), which are described with dotted lines; (b) Cross-modal connections, which are illustrated by solid lines. The inference of a deep layer will be combined with all shallower sideouts (e.g., in the teacher net, $P_R^6$ will be fed to $P_R^5$, $P_R^4$, $P_R^3$, $P_R^2$ for combination). Formally,
\begin{equation}
\tilde P_R^i = \left\{ \begin{array}{l}
{\bf{w}}_{^R}^iP_R^i + \sum\limits_{k = i + 1}^K {{\bf{\tilde w}}_{R,i}^k} \tilde P_R^k,\;i = 1,...,K-1\\
P_R^i,\;\;\;\;\;\;\;\;\;\;\;\;\;\;\;\;\;\;\;\;\;\;\;\;\;\;i = K
\end{array} \right.
\end{equation}
where ${\bf{w}}_{^R}^i$ and ${\bf{\tilde w}}_{R,i}^k$ are the weights for $P_R^i$ and the side-out $\tilde P_R^k$ from the $k^{th}$ level. 
\par Similarly for the counterpart modality $\mathcal{M}_{D}^{X}$:
\begin{equation}
\tilde P_D^i = \left\{ \begin{array}{l}
{\bf{w}}_{^D}^iP_D^i + \sum\limits_{k = i + 1}^K {{\bf{\tilde w}}_{D,i}^k} \tilde P_D^k,\;i = 1,...,K-1\\
P_D^i,\;\;\;\;\;\;\;\;\;\;\;\;\;\;\;\;\;\;\;\;\;\;\;\;\;\;i = K
\end{array} \right.
\end{equation}
where $P_D^i = \delta _d^i(\phi _{D}^i)$, $\Phi  = \{\phi _{D}^i,\;i = 1,...,K\}$ denotes the ${K}$ layered representations, $\delta _d^i$ is the learned classifier, ${\bf{w}}_{^D}^i$ and ${\bf{\tilde w}}_{D,i}^k$ are the weights for $P_D^i$ and the side-outs $\tilde P_D^k$ from the deeper levels, respectively. 
\par Our scheme for learning sufficient modal-specific representations and inference from images in the modality $\mathcal{M}_{D}^{X}$ is to train the representations ${\Phi}$ and inference $P_D^i({I_d})$ such that the combined side-out $\tilde P_D^i({I_d})$ matches the one $\tilde P_R^i({I_r})$ inferred from its paired image in the modality $\mathcal{M}_{R}^{X}$. Therefore, we measure the discrepancy between the side-outs from two modalities with a suitable loss ${g}$: 
\begin{equation}
{L_{HCD}} = \sum\limits_{\{ {I_r},{I_d}\}  \in {{D}_{r,d}}} {\sum\limits_{i = 1}^K {g\left( {\tilde P_R^i({I_r}),\tilde P_D^i({I_d})} \right)}} 
\end{equation}
In our experiments, we adopt the L2 loss $g(x,y) = \left\| {x - y} \right\|_2^2$ for measuring the distance. By minimizing ${L_{HCD}}$, the student network is encouraged to learn rich feature hierarchies for inference.

\subsection{Complementarity-aware Cross-modal Fusion}
Having learned modal-specific representations from each modality, the following step is to select the complementary ones for informative multi-modal fusion. To this end, we propose the complementarity-aware cross-modal fusion (CA-Fuse) block to explicitly select cross-modal complements. Fig. 3(a) shows the architecture of the multi-modal fusion network and Fig. 3(c) exemplifies the CA-Fuse block in the $m^{th}$ level. Formally, the adapted deep features from the RGB and depth streams are denoted as $F_R^m$ and $F_D^m$, respectively. A 1$\times$1 convolutional layer, acting as a selector, is appended after $F_D^m$ to select complementary information to enhance the RGB features via a cross-modal skip connection $\tilde F_R^m = F_R^m + {\Re _1}(F_D^m)$. It suggests that the target of using ${\Re _1}( \cdot )$ to select complementary features from $F_D^m$ can be posed as approximating the residual part, i.e., $\tilde F_R^m - F_R^m$ equivalently. Such a reformulation exposes the cross-modal complements explicitly and eases the incorporation. If $F_R^m$ is competent for inference, the solver can simply adjust the residual towards zero. Otherwise, ${\Re _1}( \cdot )$ will be pushed to distill complements from $F_D^m$ to aid $F_R^m$ for better prediction. To further encourage the determination of the residual part, the enhanced features $\tilde F_R^m$ will infer saliency $\tilde P_{R - b}^m$ and then compared to the ground-truth $\bf{Y}$. In minimizing the distance $L_R^m$ between $\tilde P_{R - b}^m$ and $\bf{Y}$, $\tilde F_R^m$ as well as ${\Re _1}(F_D^m)$ will be optimized, thereby capturing the most complementary cues from the paired modality. A symmetric residual connection is also introduced from $F_R^m$ to $F_D^m$ to capture the complements from the RGB stream to enhance the depth features. Then these features across modalities are concatenated for joint prediction. 
\subsection{Progressively Top-down Cross-modal Cross-level Fusion Pattern}
The last question regarding how to fuse the cross-modal complements sufficiently is solved by a progressively top-down fusion pattern, in which the cross-modal features are selected and combined by the CA-Fuse block in each level and the incorporated multi-modal features are selectively transmitted to the adjacent shallower layer for the cross-level combination. Concretely, the RGB-D representations $\tilde F_{m,RD}^{m + 1}$, selected from the ${{m}}$+1 layer $\tilde F_{RD}^{m + 1}$ by a transition layer (detailed parameters are illustrated in Table 1), will be upsampled by a fixed de-convolutional layer and then concatenated with $\tilde F_R^m$ and $\tilde F_D^m$ as a cross-level cross-modal representation community $\tilde F_{RD}^m$, which will be responsible for the prediction of the ${m}^{th}$ CA-Fuse block by:
\begin{equation}
P_{RD}^m = \delta _{rd}^m(\tilde F_{RD}^m),
\end{equation}
where $\delta _{rd}^m$ are the parameters for fusing cross-modal cross-level features and performing joint inference. Another cross-level fusion strategy of skip-connecting the side-outs densely is also adopted in the multi-modal fusion network and implemented by the backward prediction dense-connection (BPDC) module. The combined side-out is denoted as  
\begin{equation}
\tilde P_{RD}^m = \left\{ \begin{array}{l}
{\bf{w}}_{^{RD}}^mP_{RD}^m + \sum\limits_{k = m + 1}^K {{\bf{\tilde w}}_{RD,m}^k} \tilde P_{RD}^k,\;m = 2,...,K-1\\
P_{RD}^m,\;\;\;\;\;\;\;\;\;\;\;\;\;\;\;\;\;\;\;\;\;\;\;\;\;\;m = K
\end{array} \right.
\end{equation}
where ${\bf{w}}_{^{RD}}^m$ and ${\bf{\tilde w}}_{RD,m}^k$ denotes the weights for the predictions from the current layer and all deeper layers, respectively. The joint loss function for the multi-modal fusion network consists of the side loss from each CA-Fuse block. We also add a collaborative loss to encourage a informative combination of all the side-outs:
\begin{equation}
\begin{array}{l}
{L_{final}} = \sum\limits_{m = 1}^K {\left( {d(\tilde P_{R - b}^m,{\bf{Y}}) + d(\tilde P_{D - b}^m,{\bf{Y}}) + d(\tilde P_{RD}^m,{\bf{Y}})} \right)} \\
\;\;\;\;\;\;\;\;\; + \;d(\sum\limits_{m = 1}^K {{{{\bf{\tilde w}}}^m}} \tilde P_{RD}^m,{\bf{Y}}),
\end{array}
\end{equation}
where ${d}$ is an appropriate loss function, $\bf{Y}$ is the ground-truth mask, ${{\bf{\tilde w}}^m}$ is the weight for $\tilde P_{RD}^m$, $\tilde P_{R - b}^m$ and $\tilde P_{D - b}^m$ denote the predictions by $\tilde F_R^m$ and $\tilde F_D^m$ in the CA-Fuse block, respectively. This joint loss enables the cross-modal and cross-level combinations to be complementary for better inference.

\begin{figure*}[!t]
	\centering
	\includegraphics[height=0.37\textwidth, width=1\textwidth]{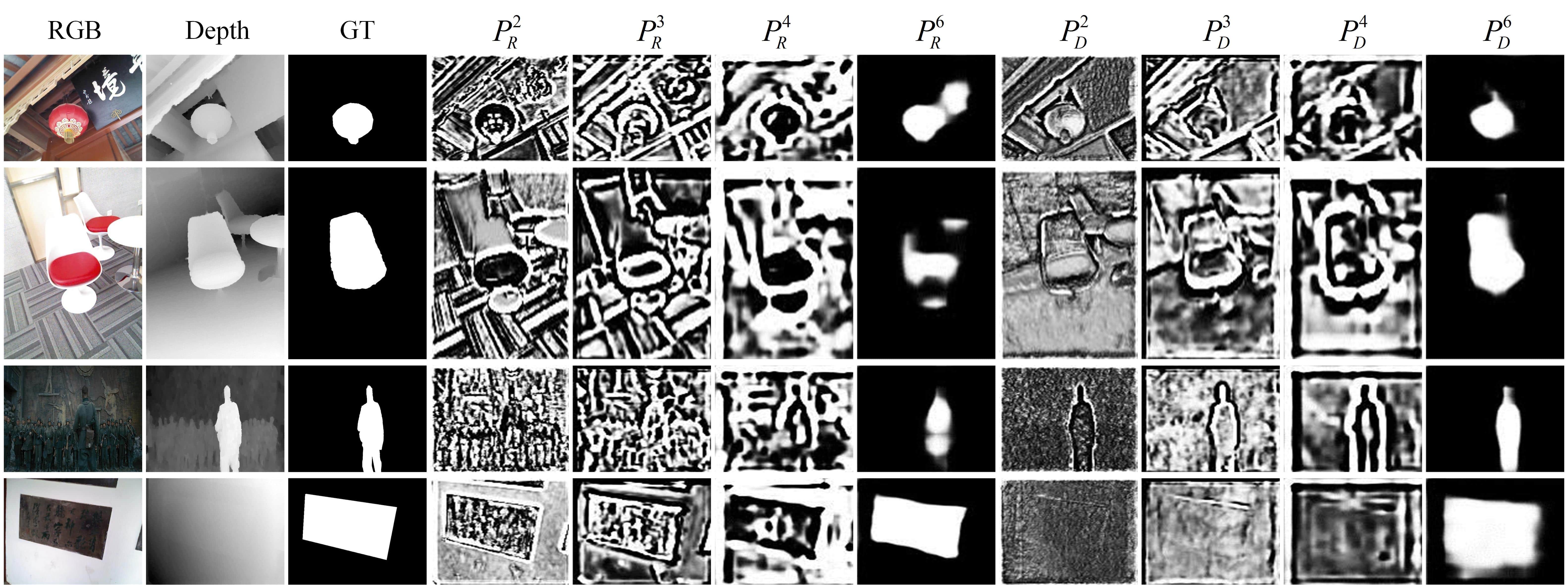}
	\setlength{\belowcaptionskip}{1pt}
	\caption{The individual inference from the teacher and student streams in the hierarchical cross-modal distillation network.}
	\label{fig4}
\end{figure*}

\section{Experiments}
In this section, we will introduce the implementation details, experimental comparisons and ablation studies to verify the advantages of the proposed method to learn, select and fuse cross-modal complements and the promise in zero-shot saliency detection from depth images.  
\subsection{Dataset and Evaluation Metrics}
We evaluate our model on three RGB-D benchmark datasets: \textbf{NLPR} \cite{4}includes 1000 indoor/outdoor RGB-D pairs collected using Kinect;  \textbf{NJUD} \cite{17} and \textbf{STEREO} \cite{20} datasets contains 2003 and 797 stereoscopic images respectively, which are generated from the Internet and 3D movies and an optical method is adopted to compute depth images. We follow the previous works \cite{12,23,29} to randomly pick 650 and 1400 RGB-D pairs from the NLPR and NJUD datasets respectively and combine them as the training set. We adopt the Precision-Recall (PR) curve, the F-measure and the mean absolute error (MAE) scores as evaluation metrics. Concretely, each saliency map $S$ will be binarized by a threshold. The converted binary mask will be compared to the ground-truth $G$ to compute the precision and recall. By varying the threshold from 0 to 255, we can obtain a series of precision-recall pairs, thereby forming the PR curve. The formulation of the F-measure is
\begin{equation}
F_\beta ^{} = \frac{{\left({1 + {\beta ^2}} \right) \cdot Precision \cdot Recall}}{{{\beta ^2} \cdot Precision + Recall}},
\end{equation}
where ${\beta ^2} = 0.3$ as suggested by \cite{9,10}.
The saliency map and binary ground-truth are normalized to [0, 1] and the MAE is to measure the pixel-wise discrepancy between the saliency map $\bar S$ and the ground-truth mask $\bar G$ averagely:
\begin{equation}
{\rm{MAE}} = \frac{1}{{W \times H}}\sum\limits_{i = 1}^W {\sum\limits_{j = 1}^H {\left| {\bar S(i,j) - \bar G(i,j)} \right|}},
\end{equation}
where ${W}$ and ${H}$ are the width and height of the saliency map.


\subsection{Implementation Details}
We conduct our experiments using the Caffe \cite{41} toolbox on a workstation with two GTX 1070 GPUs. The learning rate for the teacher network, the hierarchical cross-modal distillation network and the final RGB-D salient object detection network are $1\times10^{-7}$, $1\times10^{-6}$ and $2\times10^{-9}$, respectively. 

\renewcommand\arraystretch{1.5}
\begin{table}[]
	\caption{The Parameters of the Adaptation Layers and the Cross-level Transition Layers}
	\begin{tabular}{p{1.5cm}<{\centering}p{1.9cm}<{\centering}p{1.9cm}<{\centering}p{1.9cm}<{\centering}}
		\toprule[1pt]
		\textbf{Level}                                                 & \begin{tabular}[c]{@{}c@{}}\textbf{Adaptation 1}\end{tabular} & \begin{tabular}[c]{@{}c@{}}\textbf{Adaptation 2}\end{tabular} & \begin{tabular}[c]{@{}c@{}}\textbf{Transition layer}\end{tabular} \\ \hline
		\begin{tabular}[c]{@{}c@{}}CA-Fuse   6\end{tabular} & ---                                                       & ---                                                       & 384, 1$\times$1                                                     \\
		\begin{tabular}[c]{@{}c@{}}CA-Fuse   5\end{tabular} & 384, 1$\times$1                                                 & ---                                                      & 384, 1$\times$1                                                     \\ 
		\begin{tabular}[c]{@{}c@{}}CA-Fuse   4\end{tabular} & 384, 3$\times$3                                                 & 384, 3$\times$3                                                 & 256, 1$\times$1                                                     \\ 
		\begin{tabular}[c]{@{}c@{}}CA-Fuse   3\end{tabular} & 192, 3$\times$3                                                 & 192, 3$\times$3                                                 & 128, 1$\times$1                                                     \\ 
		\begin{tabular}[c]{@{}c@{}}CA-Fuse   2\end{tabular} & 128, 3$\times$3                                                 & 128, 3$\times$3                                                 & ---                                                           \\ \bottomrule[1pt]
	\end{tabular}
\end{table}

\par For a fair comparison with the previous works based on the VGG network, we also adopt the VGG model as the backbone for both modalities and the detailed hierarchical cross-modal transfer architecture is illustrated in Fig. 2. The trunk inherits five convolutional blocks from the original VGG model. We add a new 512 13$\times$13 convolutional layer for perceiving precedent features globally to enhance the localization ability. Then the strategy similar to \cite{42} is leveraged to generate side outputs for each level. Specifically, the last layer in each convolutional block (e.g., Conv4${\_3}$ and Conv2$\_2$) will be appended with one or two adaptation layers to the backbone. The details of the adaptation layers are shown in Table 1. We firstly train the teacher network with the architecture shown in the left of Fig.2. Concretely, the adapted features are used to infer level-specific saliency $P_R^i$ via a 1$\times$1 convolutional layer. Considering that it may be difficult for the first convolutional block to provide reliable cues, we do not involve it into inference for the teacher, student and the final RGB-D fusion network. Following Eq. (1), $P_R^i$ will be combined with the predictions from deeper layers to generate the side-out $\tilde P_R^i$ (refer to the BPDC module in Fig. 3 for implementation details). Accordingly, the loss function for the teacher network consists of the distance between the ground-truth mask and each side-out as well as the joint prediction combining all the side-outs as another constraint term:
\begin{equation}
{L_{Teac}} = \sum\limits_{i = 2}^K {d(\tilde P_R^i,{\bf{Y}})}  + \;d(\sum\limits_{i = 2}^6 {{\bf{\tilde w}}_R^i} \tilde P_R^i,{\bf{Y}}),
\end{equation}
where ${\bf{\tilde w}}_R^i$ is the weight for $\tilde P_R^i$.
\par The architecture of the student stream inherits the one of the teacher stream. When training the hierarchical cross-modal distillation network, the teacher stream is frozen. We adopt the cross-entropy loss for optimization when training the teacher network and the RGB-D fusion network:
\begin{equation}
d(\tilde P,{\bf{Y}}) = {\bf{Y}}\log \tilde P + (1 - {\bf{Y}})\log (1 - \tilde P)
\end{equation}

\subsection{On the Hierarchical Cross-modal Distillation Schema}
\subsubsection{Does the Student Network Learn Specific Cues?}
The first question we want to investigate is whether the proposed hierarchical cross-modal distillation scheme can encourage the student network to learn specific cues to complement the source modality. Fig. 4 shows the individual inference from each level without combining with the predictions from deeper layers. It is not difficult to note that the side-outs from the student network present different patterns in contrast to the ones generated by the teacher. More specifically, the shallow layers of the student network are only responsive to the depth variations while insensitive to the color/texture changes and the deeper layers, which are more responsible for locating the salient object, pay more attention to the object with distinguished depth. These differences demonstrate that the student network explores depth-specific saliency cues in each level effectively, which are complementary to the ones from the paired modality. Also, the cross-modal complementarity resides in multiple levels. These observations verify our motivations that a selector is in demand for highlighting the real complementary cues. Besides, a sufficient fusion scheme is also necessary considering the cross-modal complementarity in multiple layers.

\renewcommand\arraystretch{1.5}
\begin{table}[]
	\caption{The Performance of Zero-shot Saliency Detection from Depth Images}
	\begin{tabular}{p{0.9cm}<{\centering}p{0.8cm}<{\centering}p{0.8cm}<{\centering}p{0.9cm}<{\centering}p{0.8cm}<{\centering}p{0.8cm}<{\centering}p{0.9cm}<{\centering}}
		\toprule[1pt]
		\multirow{2}{*}{\textbf{Scheme}} & \multicolumn{3}{c}{\bm{$F_\beta$}} & \multicolumn{3}{c}{\textbf{MAE}} \\ \cline{2-7} 
		& NLPR  & NJUD  & STEREO & NLPR   & NJUD  & STEREO \\
		\textbf{A}                       & 0.705 & 0.753 & 0.733  & 0.096  & 0.131 & 0.142  \\
		\textbf{B}                       & 0.511 & 0.598 & 0.576  & 0.158  & 0.188 & 0.192  \\ 
		\textbf{C}                       & 0.747 & 0.790 &  -  & 0.088  & 0.109 & -  \\ \bottomrule[1pt]
	\end{tabular}
\end{table}


\renewcommand\arraystretch{1.5}
\begin{table}[]
	\caption{Comparisons of Different Initialization Schemes for the Depth-induced Saliency Detection and RGB-D Saliency Detection Networks}
	\label{Table3}
	\begin{tabular}{p{1.66cm}<{\centering}p{0.7cm}<{\centering}p{0.7cm}<{\centering}p{0.74cm}<{\centering}p{0.7cm}<{\centering}p{0.7cm}<{\centering}p{0.74cm}<{\centering}}
		\toprule[1pt]
		\multirow{2}{*}{\textbf{Scheme}} & \multicolumn{3}{c|}{\bm{$F_\beta$}} & \multicolumn{3}{c}{\textbf{MAE}} \\ \cline{2-7} 
		& NLPR & NJUD & \multicolumn{1}{c|}{STEREO} & NLPR & NJUD & STEREO \\ 
		\textbf{D-(A)} & \multicolumn{6}{c}{Fail to converge} \\
		\textbf{\underline{D-(B)}} & \underline{0.747} & \underline{0.780} & \multicolumn{1}{c|}{\underline{0.732}} & \underline{0.080} & \underline{0.096} & \underline{0.127} \\
		\textbf{D-(C)} & 0.784 & 0.796 & \multicolumn{1}{c|}{0.739} & 0.069 & 0.089 & 0.122 \\ 
		\textbf{$fine$(D-(C))}& \textbf{0.792} & \textbf{0.807} & \multicolumn{1}{c|}{\textbf{0.737}} & \textbf{0.066} & \textbf{0.082} & \textbf{0.118} \\ 							
		\hline			
		\textbf{\underline{RD-(A)}} & \underline{0.842} & \underline{0.854} & \multicolumn{1}{c|}{\underline{0.868}} & \underline{0.054} & \underline{0.063} & \underline{0.066} \\
		\textbf{RD-(B)} & 0.861 & 0.860 & \multicolumn{1}{c|}{0.877} & 0.049 & 0.061 & 0.061 \\
		\textbf{RD-(C)} & \textbf{0.872} & \textbf{0.871} & \multicolumn{1}{c|}{\textbf{0.880}} & \textbf{0.046} & \textbf{0.057} & \textbf{0.060} \\ \bottomrule[1pt]
	\end{tabular}%
	\begin{tablenotes}
		\small
		\item Notes: 
		
		\begin{itemize}
		 \item[-]“\textbf{D-(B)}”: Depth-induced saliency detection network initialized by task-adapted ImageNet pre-trained weights . 
		 \item[-]“\textbf{$fine$(D-(C))}”: Depth-induced saliency detection network initialized by the proposed cross-modal distillation scheme. 
		 \item[-]“\textbf{RD-(A)}”: RGB-D saliency detection network initialized by ImageNet pre-trained weights. 
		 \item[-]“\textbf{RD-(C)}”: RGB-D saliency detection network initialized by the proposed cross-modal distillation scheme. 
		\end{itemize}
		The comparison between them also demonstrates the notable improvement benefited from the distillation scheme. More details please refer to Section 4.3.3.
	\end{tablenotes}
\end{table}

\begin{figure}[!t]
	\centering
	\includegraphics[height=0.45\textwidth, width=0.48\textwidth]{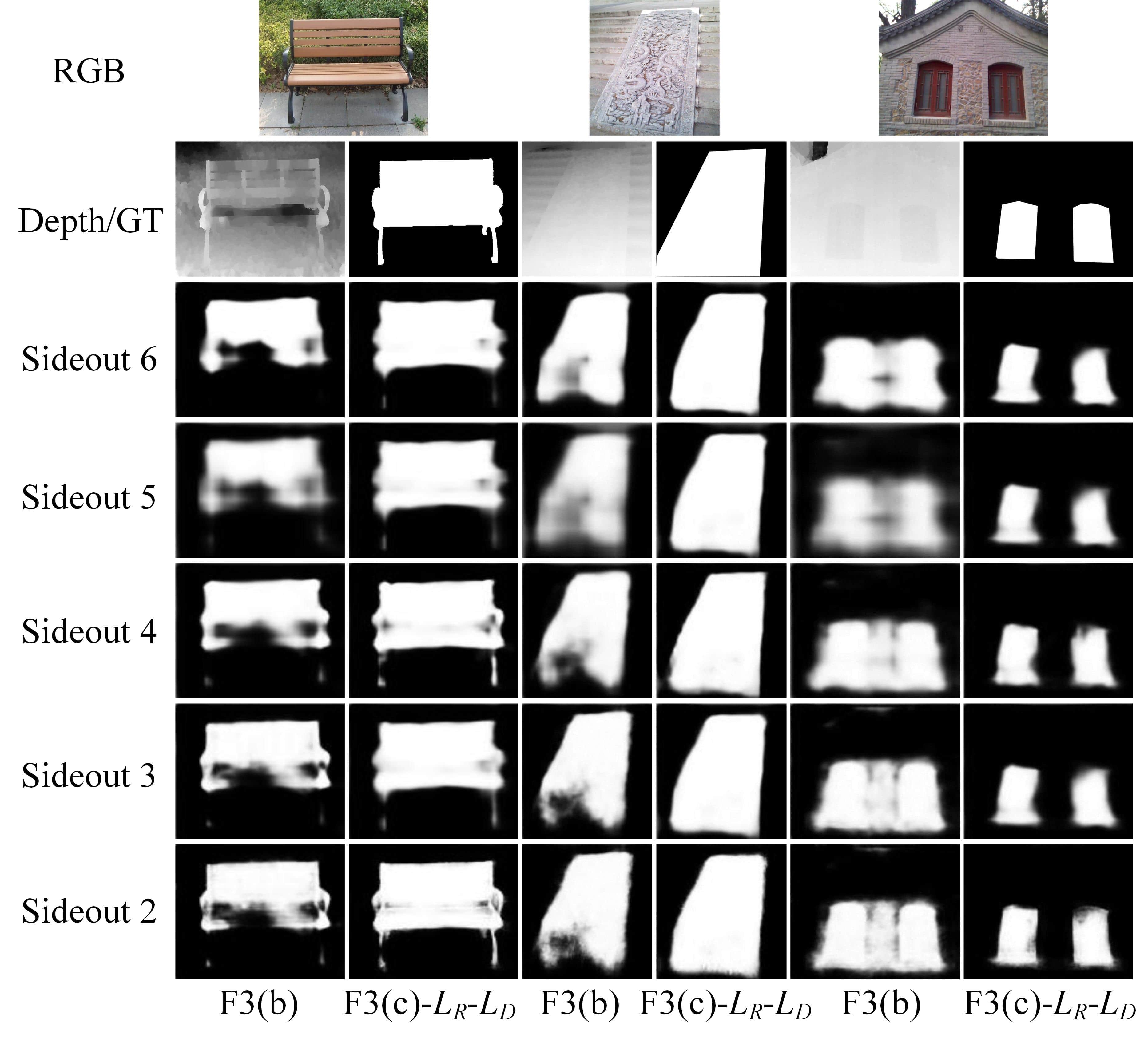}
	\setlength{\belowcaptionskip}{1pt}
	\caption{Analyze the components in the CA-Fuse block visually.}
	\label{fig5}
\end{figure}

\subsubsection{For Zero-shot Saliency Detection from Depth Images}
Our cross-modal transfer learning scheme allows zero-shot saliency detection for new modalities (e.g., depth). To verify this advantage, we combine the RGB salient object detection datasets including MSRA10K \cite{9}, ECSSD \cite{44} and SED2 \cite{45} to train the teacher network. Then we only use the RGB-D pairs from the RGB-D salient object detection training set to train the hierarchical cross-modal distillation network. We test the transferred student network with HHA (depth) images (noted as ``A"). We also test the teacher network with the HHA images as inputs for comparison (noted as ``B"). As shown in Table 2, the significant outperformance of the scheme ``A" than ``B" verifies the cross-modal discrepancy and denotes the success of the proposed cross-modal transfer method. It is able to encourage modal-specific representations and inference, offering the promise in zero-shot saliency detection for new modalities. 
\par We also report the RGB-only saliency results on the target dataset as another baseline (using RGB images as inputs to feed the teacher network, denoted as ‘C’). The RGB detector trained on the source dataset obtains satisfactory performance on the RGB images, while our zero-shot saliency detector on depth images achieve comparable detection performance (noted as ‘A’). Note that the STEREO dataset contains some same images in the source RGB saliency dataset. So we do not list the comparison results on this dataset.
\subsubsection{Advantages as A Pre-training Scheme}
In this section, we report the advantages of the proposed hierarchical cross-modal distillation scheme as a pre-training method for depth-induced saliency detection. We involve other two initialization strategies for comparison: 1) \textbf{D-(A)}: Random initialization; 2) \textbf{D-(B)}: Using the weights of the RGB-induced saliency detection network as initialization; 3) \textbf{D-(C)}: Using the proposed hierarchical cross-modal distillation schema. As shown in Table 3, we find that with ground-truth masks in each level as supervision, the convergence cannot be guaranteed when training the depth CNN from scratch, even we carefully tune the learning parameters. Compared to finetuning the RGB CNN, a huge improvement can be observed by using the hierarchical cross-modal distillation scheme, which demonstrates the efficacy of our cross-modal adaptation strategy. This improvement indicates that the side-outs from the teacher network serve as better guidance than the more correct ground-truth. We attribute this superiority to the progressive enhancement across the side-outs from the teacher. Compared to supervising each level with the same ground-truth, these side-outs become more direct and level-specific supervisions. The evolution of the side-outs demystifies level-specific contributions and cross-level collaborations explicitly. As a result, the goal of each level of the student is simplified as mimicking level-specific inference. For example, the goal of shallow layers is to learn low-level features for identifying object edges, which is a much easier task for them than predicting the completed saliency map. Additionally, finetuning the adapted student stream with ground-truth masks (denoted as ``$fine$(\textbf{D-(C)})" allows further improvement.  
\par We also report the performance of using the proposed cross-modal transfer learning method as pre-training for the final RGB-D salient object detection network. We involve other two strategies for comparison. \textbf{RD-(A)}: Both the RGB and the depth streams are initialized by the VGG model without respective finetuning with the RGB-D saliency datasets. This strategy is adopted in \cite{12}; \textbf{RD-(B)}: Stage-wise training. It means finetuning the RGB stream with the VGG model as initialization firstly. Then we train the depth stream starting from the well-trained RGB weights. This strategy is widely adopted in the previous works such as \cite{23,29}; \textbf{RD-(C)}: Using the trained hierarchical cross-modal distillation network as initialization. With the three initialization schemes, we then train the RGB-D fusion network using the RGB-D pairs and ground-truth. The comparison in Table 3 showcases the outperformance of the proposed cross-modal transfer schema, suggesting its success in learning better modal-specific representations. 
\renewcommand\arraystretch{1.5}
\begin{table}[]
	\caption{Analyze the Components in the CA-Fuse Block Quantitatively}
	\begin{tabular}{p{1.66cm}<{\centering}p{0.7cm}<{\centering}p{0.7cm}<{\centering}p{0.74cm}<{\centering}p{0.7cm}<{\centering}p{0.7cm}<{\centering}p{0.74cm}<{\centering}}
		\toprule[1pt]
		\multirow{2}{*}{\textbf{Block}} & \multicolumn{3}{c}{\bm{$F_\beta$}} & \multicolumn{3}{c}{\textbf{MAE}} \\ \cline{2-7} 
		& NLPR  & NJUD  & STEREO & NLPR   & NJUD  & STEREO \\
		\textbf{F3(b)}                  & 0.842    & 0.843    & 0.860     & 0.056    & 0.065    & 0.070   \\
		\textbf{F3(c)-$L_R$-$L_D$}            & 0.867    & 0.866    & 0.878     & 0.048    & 0.059    & 0.062   \\
		\textbf{F3(c)}                  &\textbf {0.872}    & \textbf{0.871}    & \textbf{0.880}     & \textbf{0.046}    & \textbf{0.057}    & \textbf{0.060}   \\
		\textbf{F3(c)-$\tilde F_{m,RD}^{m + 1}$}           & 0.867    & 0.862    & 0.878     & 0.047    & 0.059    & 0.060   \\ 
		\bottomrule[1pt]
	\end{tabular}
\end{table}
\begin{figure}[!t]
	\centering
	\includegraphics[height=0.33\textwidth, width=0.48\textwidth]{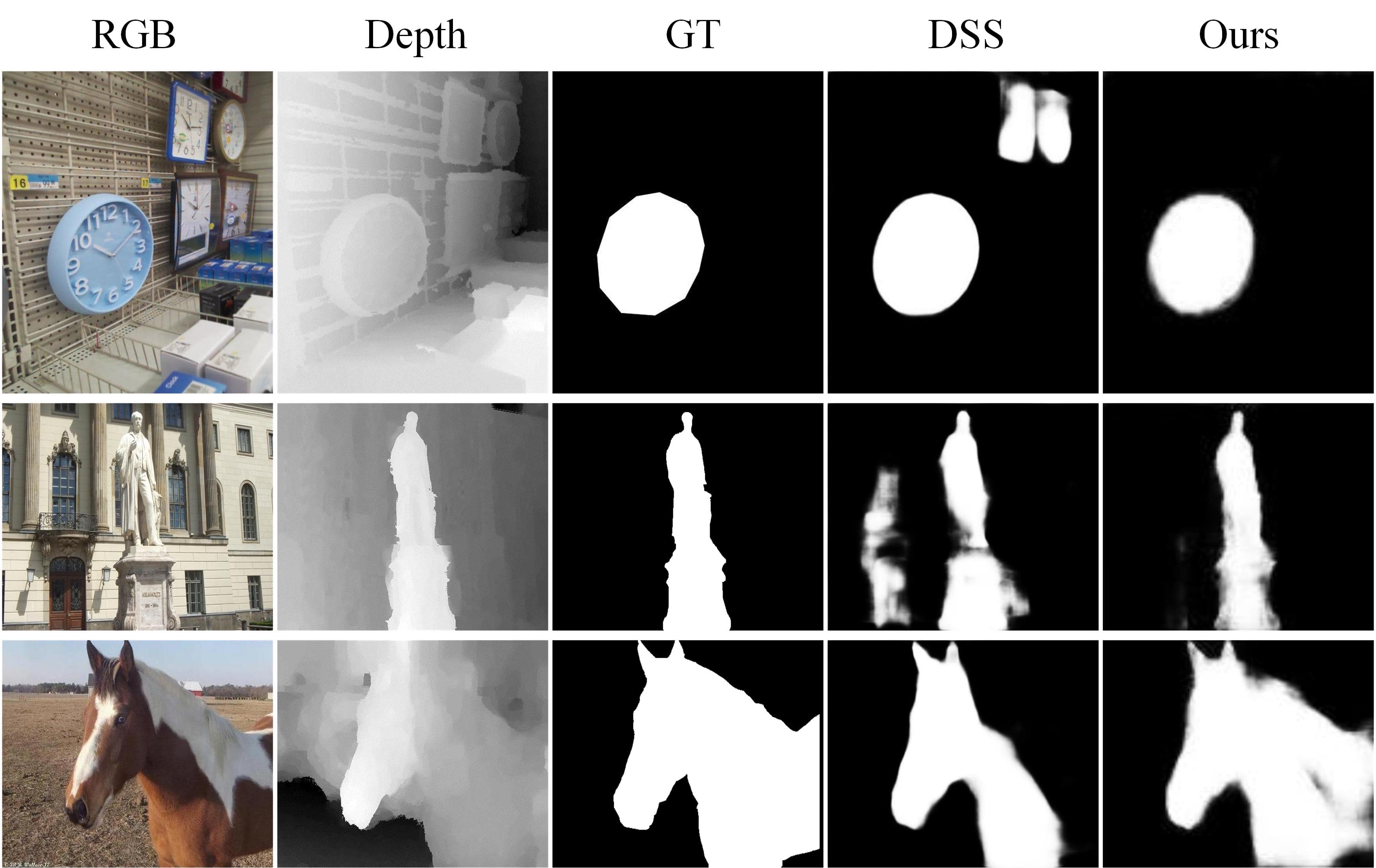}
	\setlength{\belowcaptionskip}{1pt}
	\caption{Visual comparison to state-of-the-art RGB salient object detection method.}
	\label{fig6}
\end{figure}
\begin{figure*}[!t]
	\centering
	\includegraphics[height=0.7\textwidth, width=1\textwidth]{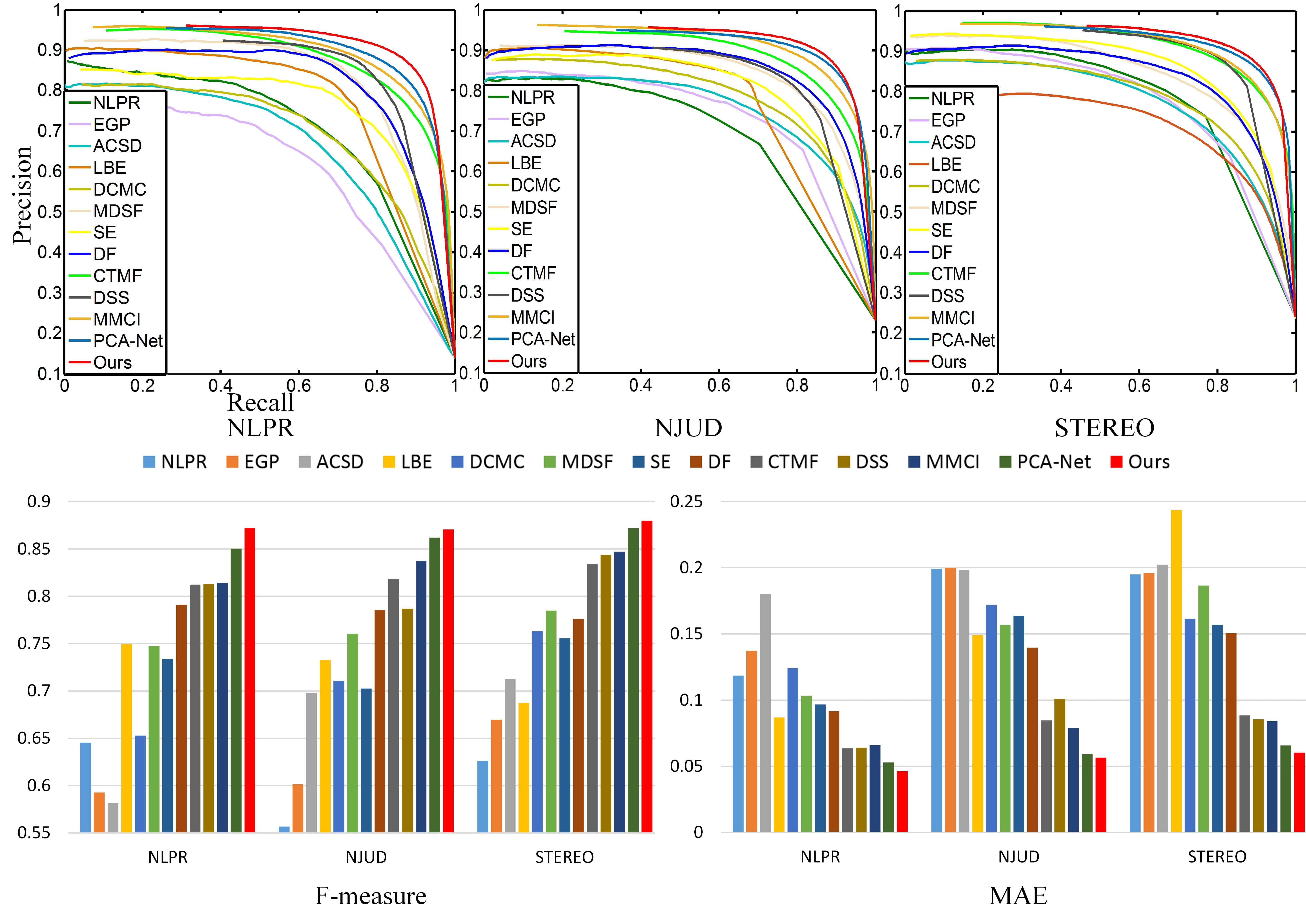}
	\setlength{\belowcaptionskip}{1pt}
	\caption{Quantitative comparison to state-of-the-art RGB and RGB-D salient object detection methods. The “MDSF” only reports the results on the NLPR and NJUD datasets.}
	\label{fig7}
\end{figure*}
\begin{figure*}[!t]
	\centering
	\includegraphics[height=0.6\textwidth, width=1\textwidth]{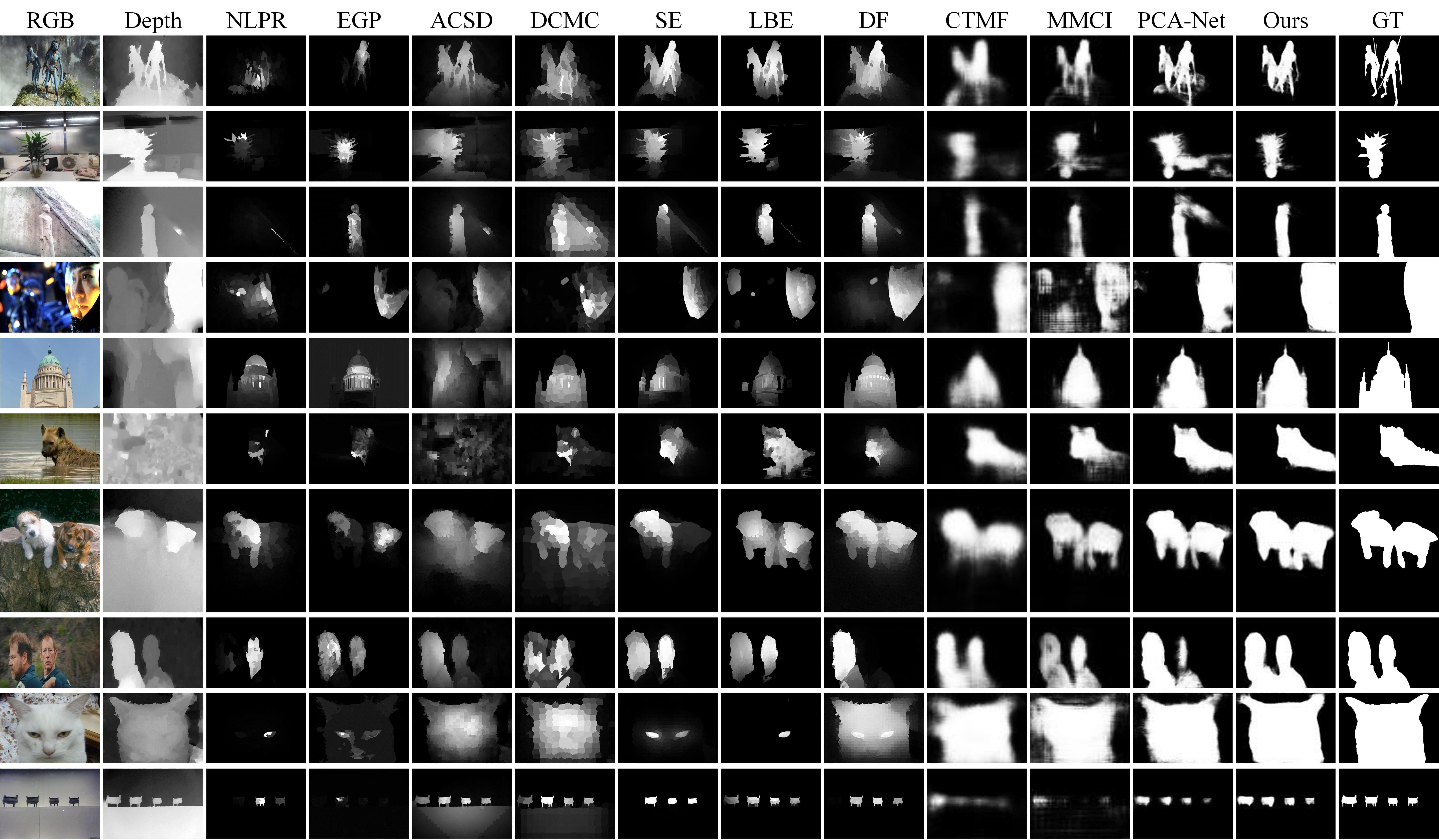}
	\setlength{\belowcaptionskip}{1pt}
	\caption{Visual comparison to state-of-the-art RGB-D salient object detection methods. }
	\label{fig8}
\end{figure*}
\subsection{On the CA-Fuse Block}
In this section, we analyze the components in the CA-Fuse block. We first study the importance of introducing cross-modal residual functions. Fig. 5 illustrates the side outputs from each level with different designs shown in Fig. 3. The columns indexed as ``F3(b)" show that the saliency maps inferred in the top-down pattern can be basically refined from coarse to fine with the help of the added supervision in each level and the cross-level combinations. However, the salient objects are not uniformly highlighted and some background regions are failed to be identified, suggesting that directly concatenating cross-modal features is incapable of capturing the complementary information sufficiently. Then we improve the ``F3(b)" block by adding cross-modal residual connections and this variant is denoted as ``F3(c)-$L_R$-$L_D$". Benefiting from the cross-modal residual functions, the complementary cues from both modalities are incorporated more easily, resulting in more informative multi-modal fusion. The comparison between ``F3(b)" and ``F3(c)-$L_R$-$L_D$" in Table 4 verifies the large performance gains from the cross-modal residual connections. Moreover, the comparison between ``F3(c)-$L_R$-$L_D$" and ``F3(c)" verifies the benefits of adding supervisions on the RGB and depth branches ($L_R$ and $L_D$), which further boost the emergence of the complementary cues from the paired modality. Another question we want to study is whether it is beneficial to transmit cross-level features to the adjacent shallower layer? To answer this question, we remove the $\tilde F_{m,RD}^{m + 1}$ in Fig. 3(c). Accordingly, $\tilde F_R^m$ and $\tilde F_D^m$ will be concatenated for joint inference. We denote this variant as ``F3(c)-$\tilde F_{m,RD}^{m + 1}$". The quantitative comparison in Table 4 reports the noticeable gains by transmitting the cross-level features. We attribute this improvement to the richer RGB-D representations due to combining cross-level features.

\subsection{Comparison to State-of-the-art Methods}
We compare our model (the variant ``RD-(C)" ) to 11 state-of-the-art RGB-D salient object detection methods: NLPR \cite{4}, EGP \cite{46}, ACSD \cite{17}, DCMC \cite{18}, LBE \cite{13}, MDSF \cite{19}, SE \cite{47}, DF \cite{22}, CTMF \cite{23}, MMCI \cite{29} and our preliminary work PCA-Net \cite{12}, among which DF \cite{22}, CTMF \cite{23}, MMCI \cite{29} and PCA-Net \cite{12} are CNN-based methods. We also compare our method with a state-of-the-art RGB salient object detection model DSS \cite{42} to verify the benefits of the synchronized depth data. Fig. 6 presents the comparison to the RGB-induced saliency visually. It can be noted that when the salient object and the background are with similar appearance or the background is seriously cluttered or the salient object is non-uniform, it is difficult to locate the salient object correctly and highlight the salient regions uniformly by relying on RGB inputs only. In these scenes, our model effectively incorporates the complementary cues from the paired depth data to overcome these deficiencies to identify the real salient object and highlight the salient regions consistently. The quantitative comparison in Fig. 7 shows that our proposed method outperforms others significantly. Compared to other RGB-D salient object detection methods, the proposed one holds distinguished advantages in learning, selecting and fusing cross-modal complements. The methods \cite{4,46,47,13,17,18,19} based on handcrafted RGB-D features are easy to be confused by complex background and intra-variant salient objects due to the lack of high-level global contexts. Previous CNN-based methods \cite{22,23} that combine cross-modal features only in a single level are incapable of capturing the cross-modal complementarity residing in high-level contexts and low-level spatial cues simultaneously. The ``early fusion" schema adopted in \cite{22} results in inconsistent highlighting of salient regions and the ``late fusion" strategy used in \cite{23} leads to severely blurred saliency maps. Although the work \cite{29} remedies this shortcoming by designing two branches for global reasoning and local capturing respectively, it only leverages the last fully connected layer and an intermediate convolutional layer for joint inference. The final results are combined by directly summing the results from two branches, which is unlikely to combine local spatial cues and global contexts robustly. In contrast, our preliminary work \cite{12} involves the information in all layers via a top-down path, which is able to progressively select and fuse the complements from each modal/level and refine the saliency maps gradually. By further using the hierarchical cross-modal distillation schema proposed in this extended work, the salient object is better located. Besides, the saliency maps are more uniform and carry better details than the ones generated by \cite{12}, implying the advantages of the proposed cross-modal transfer scheme in learning better modal-specific representations. In various challenging scenes shown in Fig. 8, such as the background is complex (the $1^{st}$-$2^{nd}$ rows); the salient object and background have indistinguishable appearance or depth (the $3^{rd}$-$4^{th}$ and $5^{th}$-$6^{th}$ rows, resp.); the appearance or depth in the salient objects is non-uniform (the $7^{th}$ and $8^{th}$ rows, resp.); large/small salient objects (the $9^{th}$ and $10^{th}$ rows, resp.); multiple separated salient objects (the $10^{th}$  row). In these cases, our proposed model can learn rich representations from each modality, select the complementary cues and fuse them informatively for successful joint inference. 

\section{Conclusion}
In this paper, we propose a comprehensive view and a systematic solution for RGB-D salient object detection. The philosophy in designing an RGB-D system is generalized as three keys: modal-specific representations learning, complementary information selection and cross-modal complements fusion. Accordingly, we propose a new cross-modal transfer learning scheme, an explicit cross-modal complementarity selector and a sufficient cross-modal cross-level fusion pattern. The proposed solution solves the problems of zero-shot detection and multi-modal fusion jointly. We believe the insights provided from this work will allow us to learn better representations from new unlabeled modalities and more sufficient fusion for other multi-modal systems. 
\ifCLASSOPTIONcompsoc
  \section*{Acknowledgments}
\else
  \section*{Acknowledgment}
\fi

This work was supported by the Research Grants Council of Hong Kong (Project No CityU 11205015 and CityU 11255716).

\ifCLASSOPTIONcaptionsoff
  \newpage
\fi



\bibliographystyle{IEEEtran}
\bibliography{IEEEabrv,ref_TPAMI_update}

\begin{thebibliography}{10}
\providecommand{\url}[1]{#1}
\csname url@samestyle\endcsname
\providecommand{\newblock}{\relax}
\providecommand{\bibinfo}[2]{#2}
\providecommand{\BIBentrySTDinterwordspacing}{\spaceskip=0pt\relax}
\providecommand{\BIBentryALTinterwordstretchfactor}{4}
\providecommand{\BIBentryALTinterwordspacing}{\spaceskip=\fontdimen2\font plus
\BIBentryALTinterwordstretchfactor\fontdimen3\font minus
  \fontdimen4\font\relax}
\providecommand{\BIBforeignlanguage}[2]{{%
\expandafter\ifx\csname l@#1\endcsname\relax
\typeout{** WARNING: IEEEtran.bst: No hyphenation pattern has been}%
\typeout{** loaded for the language `#1'. Using the pattern for}%
\typeout{** the default language instead.}%
\else
\language=\csname l@#1\endcsname
\fi
#2}}
\providecommand{\BIBdecl}{\relax}
\BIBdecl

\bibitem{1}
J.~Han, L.~Shao, D.~Xu, and J.~Shotton, ``Enhanced computer vision with
  microsoft kinect sensor: A review,'' \emph{IEEE Trans. Cybern.}, vol.~43,
  no.~5, pp. 1318--1334, 2013.

\bibitem{2}
S.~Gupta, P.~Arbel{\'a}ez, R.~Girshick, and J.~Malik, ``Indoor scene
  understanding with rgb-d images: Bottom-up segmentation, object detection and
  semantic segmentation,'' \emph{Int. J. Comput. Vis.}, vol. 112, no.~2, pp.
  133--149, 2015.

\bibitem{3}
M.~Camplani, S.~L. Hannuna, M.~Mirmehdi, D.~Damen, A.~Paiement, L.~Tao, and
  T.~Burghardt, ``Real-time rgb-d tracking with depth scaling kernelised
  correlation filters and occlusion handling.'' in \emph{Proc. British Mach.
  Vis. Conf.}, 2015, pp. 145--1.

\bibitem{4}
H.~Peng, B.~Li, W.~Xiong, W.~Hu, and R.~Ji, ``Rgbd salient object detection: a
  benchmark and algorithms,'' in \emph{Proc. Eur. Conf. Comput. Vis.}, 2014,
  pp. 92--109.

\bibitem{5}
L.~Shao and M.~Brady, ``Specific object retrieval based on salient regions,''
  \emph{Pattern Recognit.}, vol.~39, no.~10, pp. 1932--1948, 2006.

\bibitem{6}
V.~Mahadevan, N.~Vasconcelos \emph{et~al.}, ``Biologically inspired object
  tracking using center-surround saliency mechanisms.'' \emph{IEEE Trans.
  Pattern Anal. Mach. Intell.}, vol.~35, no.~3, pp. 541--554, 2013.

\bibitem{7}
J.~Harel, C.~Koch, and P.~Perona, ``Graph-based visual saliency,'' in
  \emph{Proc. Adv. Neural Inf. Process. Syst.}, 2007, pp. 545--552.

\bibitem{8}
J.~Yang and M.-H. Yang, ``Top-down visual saliency via joint crf and dictionary
  learning,'' \emph{IEEE Trans. Pattern Anal. Mach. Intell.}, vol.~39, no.~3,
  pp. 576--588, 2017.

\bibitem{9}
M.-M. Cheng, N.~J. Mitra, X.~Huang, P.~H. Torr, and S.-M. Hu, ``Global contrast
  based salient region detection,'' \emph{IEEE Trans. Pattern Anal. Mach.
  Intell.}, vol.~37, no.~3, pp. 569--582, 2015.

\bibitem{10}
A.~Borji, M.-M. Cheng, H.~Jiang, and J.~Li, ``Salient object detection: A
  benchmark,'' \emph{IEEE Trans. Image Process.}, vol.~24, no.~12, pp.
  5706--5722, 2015.

\bibitem{43}
S.~Gupta, R.~Girshick, P.~Arbel{\'a}ez, and J.~Malik, ``Learning rich features
  from rgb-d images for object detection and segmentation,'' in \emph{Proc.
  Eur. Conf. Comput. Vis.}, 2014, pp. 345--360.

\bibitem{11}
H.~Fu, D.~Xu, S.~Lin, and J.~Liu, ``Object-based rgbd image co-segmentation
  with mutex constraint,'' in \emph{Proc. IEEE Conf. Comput. Vis. Pattern
  Recog.}, 2015, pp. 4428--4436.

\bibitem{48}
H.~Fu, D.~Xu, and S.~Lin, ``Object-based multiple foreground segmentation in
  rgbd video,'' \emph{IEEE Trans. Image Process.}, vol.~26, no.~3, pp.
  1418--1427, 2017.

\bibitem{49}
R.~Cong, J.~Lei, H.~Fu, W.~Lin, Q.~Huang, X.~Cao, and C.~Hou, ``An iterative
  co-saliency framework for rgbd images,'' \emph{IEEE Trans. Cybern.}, 2017.

\bibitem{50}
R.~Cong, J.~Lei, H.~Fu, Q.~Huang, X.~Cao, and C.~Hou, ``Co-saliency detection
  for rgbd images based on multi-constraint feature matching and cross label
  propagation,'' \emph{IEEE Trans. Image Process.}, vol.~27, no.~2, pp.
  568--579, 2018.

\bibitem{12}
H.~Chen and Y.~Li, ``Progressively complementarity-aware fusion network for
  rgb-d salient object detection,'' in \emph{Proc. IEEE Conf. Comput. Vis.
  Pattern Recog.}, 2018, pp. 3051--3060.

\bibitem{13}
D.~Feng, N.~Barnes, S.~You, and C.~McCarthy, ``Local background enclosure for
  rgb-d salient object detection,'' in \emph{Proc. IEEE Conf. Comput. Vis.
  Pattern Recog.}, 2016, pp. 2343--2350.

\bibitem{14}
C.~Lang, T.~V. Nguyen, H.~Katti, K.~Yadati, M.~Kankanhalli, and S.~Yan, ``Depth
  matters: Influence of depth cues on visual saliency,'' in \emph{Proc. Eur.
  Conf. Comput. Vis.}, 2012, pp. 101--115.

\bibitem{15}
K.~Desingh, K.~M. Krishna, D.~Rajan, and C.~Jawahar, ``Depth really matters:
  Improving visual salient region detection with depth.'' in \emph{Proc.
  British Mach. Vis. Conf.}, 2013.

\bibitem{16}
A.~Ciptadi, T.~Hermans, and J.~M. Rehg, ``An in depth view of saliency,'' in
  \emph{Proc. British Mach. Vis. Conf.}, 2013.

\bibitem{17}
R.~Ju, L.~Ge, W.~Geng, T.~Ren, and G.~Wu, ``Depth saliency based on anisotropic
  center-surround difference,'' in \emph{Proc. IEEE Int. Conf. Image Process.},
  2014, pp. 1115--1119.

\bibitem{18}
R.~Cong, J.~Lei, C.~Zhang, Q.~Huang, X.~Cao, and C.~Hou, ``Saliency detection
  for stereoscopic images based on depth confidence analysis and multiple cues
  fusion,'' \emph{Signal Process. Lett.}, vol.~23, no.~6, pp. 819--823, 2016.

\bibitem{19}
H.~Song, Z.~Liu, H.~Du, G.~Sun, O.~Le~Meur, and T.~Ren, ``Depth-aware salient
  object detection and segmentation via multiscale discriminative saliency
  fusion and bootstrap learning,'' \emph{IEEE Trans. Image Process.}, vol.~26,
  no.~9, pp. 4204--4216, 2017.

\bibitem{20}
Y.~Niu, Y.~Geng, X.~Li, and F.~Liu, ``Leveraging stereopsis for saliency
  analysis,'' in \emph{Proc. IEEE Conf. Comput. Vis. Pattern Recog.}, 2012, pp.
  454--461.

\bibitem{22}
L.~Qu, S.~He, J.~Zhang, J.~Tian, Y.~Tang, and Q.~Yang, ``Rgbd salient object
  detection via deep fusion,'' \emph{IEEE Trans. Image Process.}, vol.~26,
  no.~5, pp. 2274--2285, 2017.

\bibitem{23}
J.~Han, H.~Chen, N.~Liu, C.~Yan, and X.~Li, ``Cnns-based rgb-d saliency
  detection via cross-view transfer and multiview fusion,'' \emph{IEEE Trans.
  Cybern.}, 2017.

\bibitem{21}
A.~Krizhevsky, I.~Sutskever, and G.~E. Hinton, ``Imagenet classification with
  deep convolutional neural networks,'' in \emph{Proc. Adv. Neural Inf.
  Process. Syst.}, 2012, pp. 1097--1105.

\bibitem{24}
A.~Wang, J.~Cai, J.~Lu, and T.-J. Cham, ``Mmss: Multi-modal sharable and
  specific feature learning for rgb-d object recognition,'' in \emph{Proc. IEEE
  Int. Conf. Comput. Vis.}, 2015, pp. 1125--1133.

\bibitem{25}
S.-J. Park, K.-S. Hong, and S.~Lee, ``Rdfnet: Rgb-d multi-level residual
  feature fusion for indoor semantic segmentation,'' in \emph{Proc. IEEE Int.
  Conf. Comput. Vis.}, 2017.

\bibitem{26}
X.~Xu, Y.~Li, G.~Wu, and J.~Luo, ``Multi-modal deep feature learning for rgb-d
  object detection,'' \emph{Pattern Recognit.}, vol.~72, pp. 300--313, 2017.

\bibitem{27}
H.~Zhu, J.-B. Weibel, and S.~Lu, ``Discriminative multi-modal feature fusion
  for rgbd indoor scene recognition,'' in \emph{Proc. IEEE Conf. Comput. Vis.
  Pattern Recog.}, 2016, pp. 2969--2976.

\bibitem{28}
X.~Fan, Z.~Liu, and G.~Sun, ``Salient region detection for stereoscopic
  images,'' in \emph{Proc. IEEE Int. Conf. Digital Signal Process.}, 2014, pp.
  454--458.

\bibitem{29}
H.~Chen, Y.~Li, and D.~Su, ``Multi-modal fusion network with multi-scale
  multi-path and cross-modal interactions for rgb-d salient object detection,''
  \emph{Pattern Recognit.}, 2018.

\bibitem{30}
Y.~Cheng, R.~Cai, Z.~Li, X.~Zhao, and K.~Huang, ``Localitysensitive
  deconvolution networks with gated fusion for rgb-d indoor semantic
  segmentation,'' in \emph{Proc. IEEE Conf. Comput. Vis. Pattern Recog.},
  vol.~3, 2017.

\bibitem{31}
D.~Lin, G.~Chen, D.~Cohen-Or, P.-A. Heng, and H.~Huang, ``Cascaded feature
  network for semantic segmentation of rgb-d images,'' in \emph{Proc. IEEE Int.
  Conf. Comput. Vis.}, 2017, pp. 1320--1328.

\bibitem{32}
G.~Lin, A.~Milan, C.~Shen, and I.~Reid, ``Refinenet: Multi-path refinement
  networks for high-resolution semantic segmentation,'' in \emph{Proc. IEEE
  Conf. Comput. Vis. Pattern Recog.}, vol.~1, no.~2, 2017, p.~3.

\bibitem{33}
G.~Hinton, O.~Vinyals, and J.~Dean, ``Distilling the knowledge in a neural
  network,'' \emph{arXiv preprint arXiv:1503.02531}, 2015.

\bibitem{34}
A.~Romero, N.~Ballas, S.~E. Kahou, A.~Chassang, C.~Gatta, and Y.~Bengio,
  ``Fitnets: Hints for thin deep nets,'' \emph{arXiv preprint arXiv:1412.6550},
  2014.

\bibitem{35}
Q.~Li, S.~Jin, and J.~Yan, ``Mimicking very efficient network for object
  detection,'' in \emph{Proc. IEEE Conf. Comput. Vis. Pattern Recog.}, 2017,
  pp. 7341--7349.

\bibitem{36}
Z.~Huang and N.~Wang, ``Like what you like: Knowledge distill via neuron
  selectivity transfer,'' \emph{arXiv preprint arXiv:1707.01219}, 2017.

\bibitem{51}
C.~M. Christoudias, R.~Urtasun, M.~Salzmann, and T.~Darrell, ``Learning to
  recognize objects from unseen modalities,'' in \emph{Proc. Eur. Conf. Comput.
  Vis.}, 2010, pp. 677--691.

\bibitem{52}
R.~Socher, M.~Ganjoo, C.~D. Manning, and A.~Ng, ``Zero-shot learning through
  cross-modal transfer,'' in \emph{Proc. Adv. Neural Inf. Process. Syst.},
  2013, pp. 935--943.

\bibitem{53}
J.~Ngiam, A.~Khosla, M.~Kim, J.~Nam, H.~Lee, and A.~Y. Ng, ``Multimodal deep
  learning,'' in \emph{Proc. Int. Conf. Mach. Learn.}, 2011, pp. 689--696.

\bibitem{54}
J.~Hoffman, S.~Gupta, and T.~Darrell, ``Learning with side information through
  modality hallucination,'' in \emph{Proc. IEEE Conf. Comput. Vis. Pattern
  Recog.}, 2016, pp. 826--834.

\bibitem{55}
N.~C. Garcia, P.~Morerio, and V.~Murino, ``Modality distillation with multiple
  stream networks for action recognition,'' in \emph{Proc. Eur. Conf. Comput.
  Vis.}, 2018, pp. 103--118.

\bibitem{37}
S.~Gupta, J.~Hoffman, and J.~Malik, ``Cross modal distillation for supervision
  transfer,'' in \emph{Proc. IEEE Conf. Comput. Vis. Pattern Recog.}, 2016, pp.
  2827--2836.

\bibitem{40}
K.~Lenc and A.~Vedaldi, ``Understanding image representations by measuring
  their equivariance and equivalence,'' in \emph{Proc. IEEE Conf. Comput. Vis.
  Pattern Recog.}, 2015, pp. 991--999.

\bibitem{41}
Y.~Jia, E.~Shelhamer, J.~Donahue, S.~Karayev, J.~Long, R.~Girshick,
  S.~Guadarrama, and T.~Darrell, ``Caffe: Convolutional architecture for fast
  feature embedding,'' in \emph{Proc. ACM Int. Conf. Multimedia}, 2014, pp.
  675--678.

\bibitem{42}
Q.~Hou, M.-M. Cheng, X.~Hu, A.~Borji, Z.~Tu, and P.~Torr, ``Deeply supervised
  salient object detection with short connections,'' in \emph{Proc. IEEE Conf.
  Comput. Vis. Pattern Recog.}, 2017, pp. 5300--5309.

\bibitem{44}
Q.~Yan, L.~Xu, J.~Shi, and J.~Jia, ``Hierarchical saliency detection,'' in
  \emph{Proc. IEEE Conf. Comput. Vis. Pattern Recog.}, 2013, pp. 1155--1162.

\bibitem{45}
S.~Alpert, M.~Galun, R.~Basri, and A.~Brandt, ``Image segmentation by
  probabilistic bottom-up aggregation and cue integration,'' in \emph{Proc.
  IEEE Conf. Comput. Vis. Pattern Recog.}, 2007, pp. 1--8.

\bibitem{46}
J.~Ren, X.~Gong, L.~Yu, W.~Zhou, and M.~Ying~Yang, ``Exploiting global priors
  for rgb-d saliency detection,'' in \emph{Proc. IEEE Conf. Comput. Vis.
  Pattern Recog. Worksh.}, 2015, pp. 25--32.

\bibitem{47}
J.~Guo, T.~Ren, and J.~Bei, ``Salient object detection for rgb-d image via
  saliency evolution,'' in \emph{Proc. IEEE Int. Conf. Multimedia and Expo},
  2016, pp. 1--6.

\end{thebibliography}
%




%




\end{document}